\documentclass{midl} 

\usepackage{mwe} 
\usepackage{array}
\usepackage{url}
\jmlrvolume{-- Under Review}
\jmlryear{2019}
\jmlrworkshop{Full Paper -- MIDL 2019 submission}

\title[3D multirater RCNN]{3D multirater RCNN for multimodal multiclass detection and characterisation of extremely small objects}

\midlauthor{\Name{Carole H. Sudre\nametag{$^{1,2,3}$}} \Email{carole.sudre@kcl.ac.uk}\\
\Name{Beatriz Gomez Anson\nametag{$^{4}$}} \Email{bgomeza@santpau.cat}\\
\Name{Silvia Ingala\nametag{$^{5}$}} \Email{s.ingala@vumc.nl}\\
\Name{Chris D Lane\nametag{$^{2}$}} \Email{c.lane@ucl.ac.uk}\\
\Name{Daniel Jimenez\nametag{$^{2}$}} \Email{d.jimenez@ucl.ac.uk}\\
\Name{Lukas Haider\nametag{$^{6}$}} \Email{l.haider@ucl.ac.uk}\\
\Name{Thomas Varsavsky\nametag{$^{1,3}$}} \Email{thomas.varsavsky@kcl.ac.uk}\\
\Name{Lorna Smith\nametag{$^{7}$}} \Email{lorna.smith@ucl.ac.uk}\\
\Name{Rolf H J{\"a}ger\nametag{$^{8}$}} \Email{r.jager@ucl.ac.uk}\\
\Name{M. Jorge Cardoso\nametag{$^{1,2,3}$}} \Email{m.jorge.cardoso@kcl.ac.uk}\\
\addr $^{1}$ School of Biomedical Engineering and Imaging Sciences, King's College London, UK \\
\addr $^{2}$ Dementia Research Centre, UCL Institute of Neurology, UK\\
\addr $^{3}$ Department of Medical Physics and Biomedical Engineering, University College London, UK\\
\addr $^{4}$ Santa Creu i Sant Pau Hospital, Universitat Autonòma Barcelona, Barcelona, Spain\\
\addr $^{5}$ Vrije University Medical Centre Amsterdam, The Netherlands\\
\addr $^{6}$ Queen Square Multiple Sclerosis Centre, UCL Institute of Neurology, London, UK\\
\addr $^{7}$ Cardiometabolic Phenotyping Group, Institute of Cardiovascular Science, UCL, London, UK\\
\addr $^{8}$ Brain Repair and Rehabilitation Group, Institute of Neurology, UCL, London,  UK
}
\begin{document}

\maketitle
\vskip -15pt
\begin{abstract}
Extremely small objects (ESO) have become observable on clinical routine magnetic resonance imaging acquisitions, thanks to a reduction in acquisition time at higher resolution. Despite their small size (usually $<$10 voxels per object for an image of more than $10^6$ voxels), these markers reflect tissue damage and need to be accounted for to investigate the complete phenotype of complex pathological pathways. In addition to their very small size, variability in shape and appearance leads to high labelling variability across human raters, resulting in a very noisy gold standard. 
Such objects are notably present in the context of cerebral small vessel disease where enlarged perivascular spaces and lacunes, commonly observed in the ageing population, are thought to be associated with acceleration of cognitive decline and risk of dementia onset. In this work, we redesign the RCNN model to scale to 3D data, and to jointly detect and characterise these important markers of age-related neurovascular changes. We also propose training strategies enforcing the detection of extremely small objects, ensuring a tractable and stable training process.
\vskip -5pt
\end{abstract}
\section{Introduction}
\label{sec:intro}
\begin{figure}[t!]
    \centering
    \begin{tabular}{cc}
    Lacune & EPVS \\
    \includegraphics[trim=5cm 7.4cm 6cm 5cm , clip=true, width=0.4\textwidth]{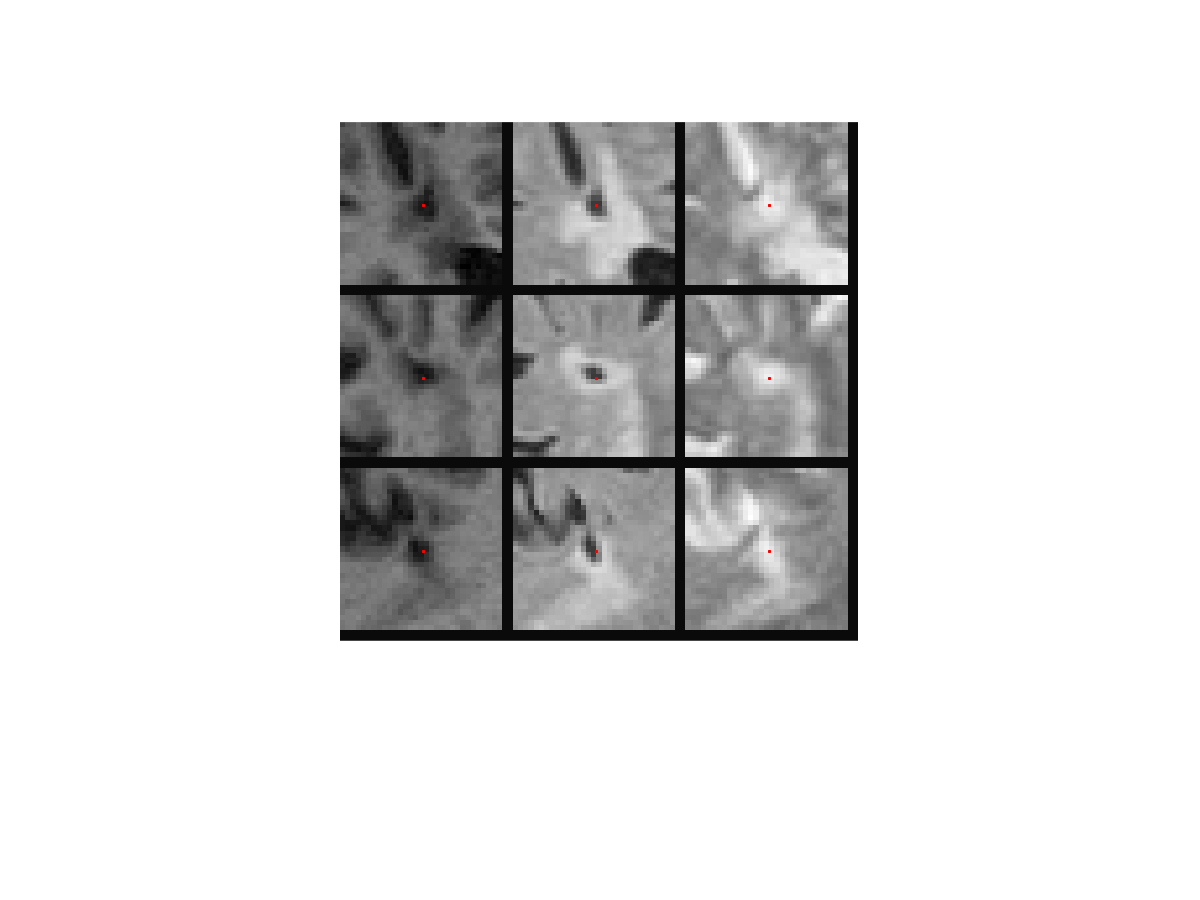} & \includegraphics[trim=5cm 7.4cm 6cm 5cm  , clip=true, width=0.4\textwidth]{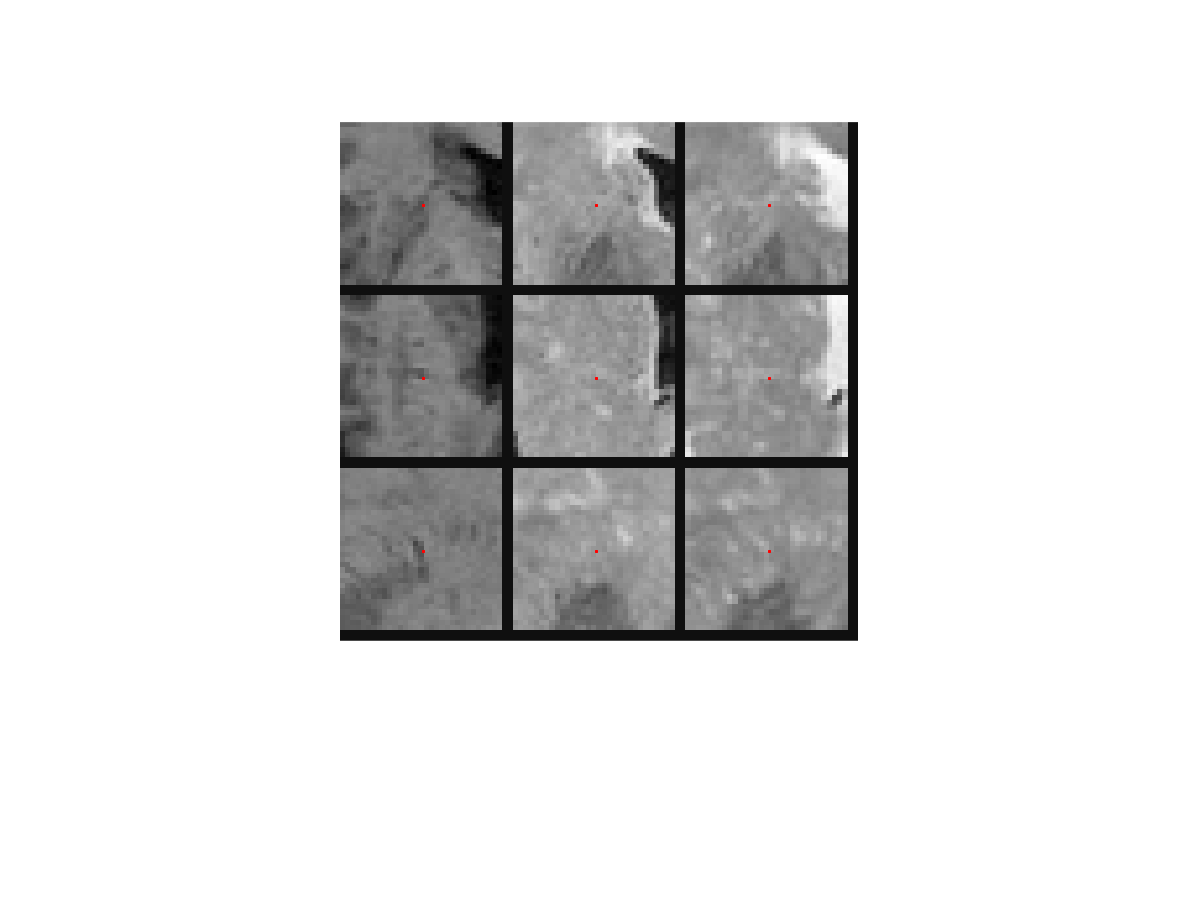}\\
    \includegraphics[trim=5cm 7.4cm 6cm 5cm , clip=true, width=0.4\textwidth]{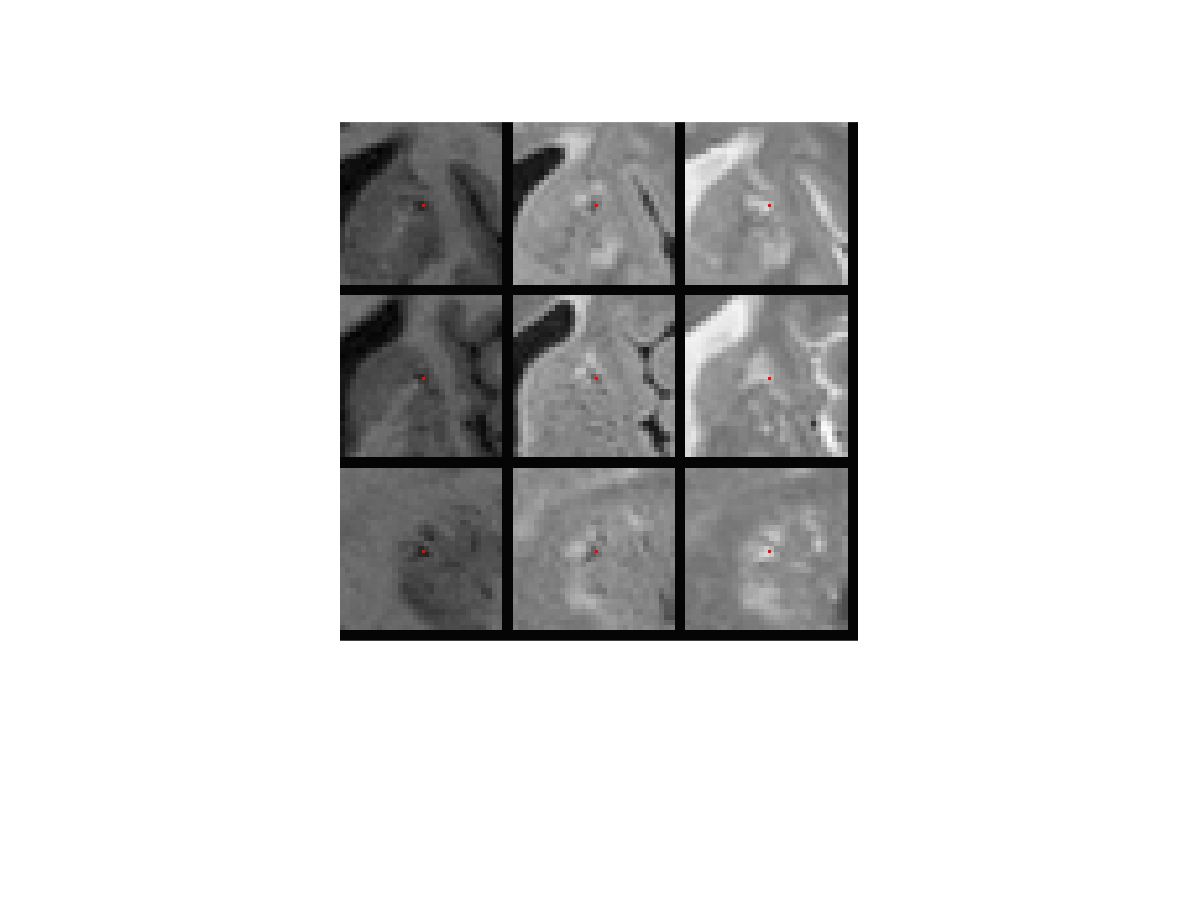} & \includegraphics[trim=5cm 7.4cm 6cm 5cm  , clip=true, width=0.4\textwidth]{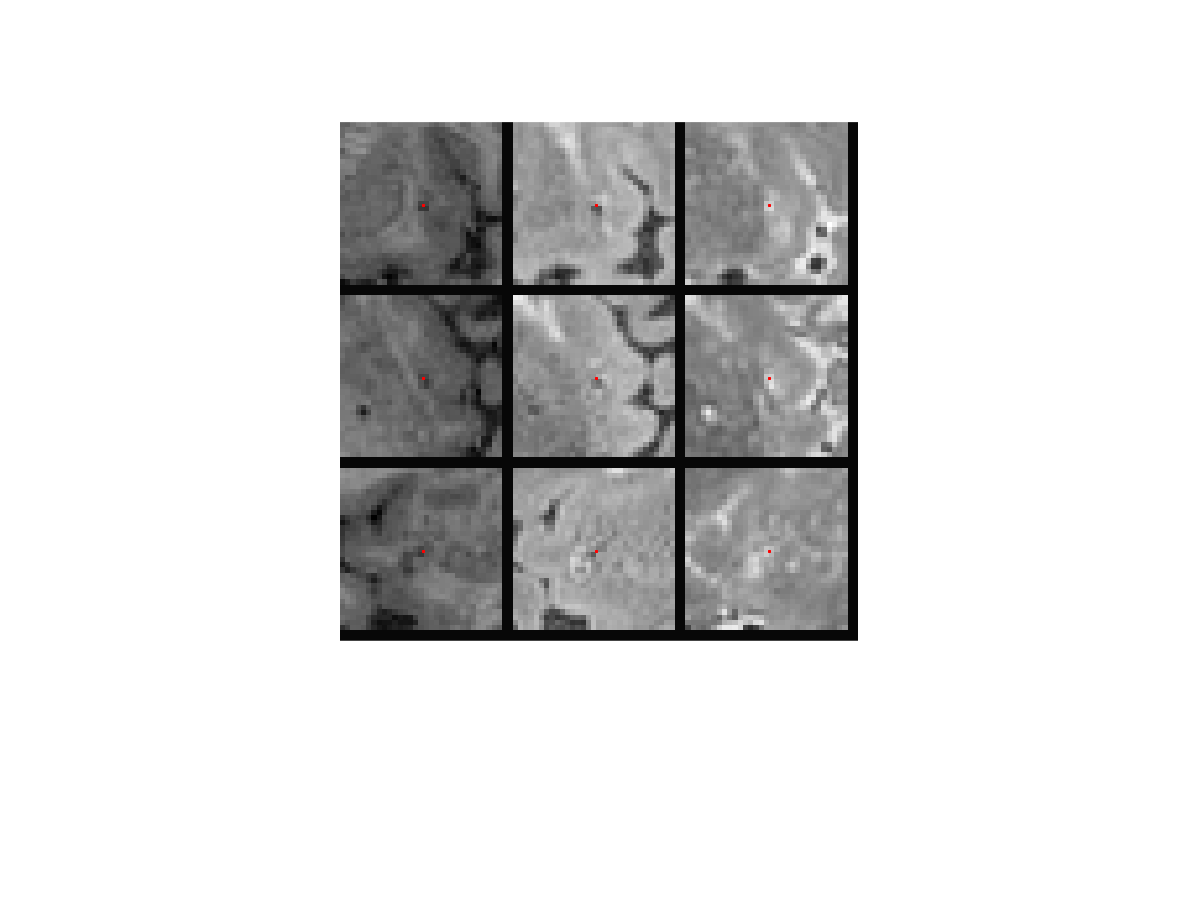}\\
    EPVS/Lacune ? & EPVS/Lacune ?\\
    \end{tabular}
    \caption{Examples of EPVS and lacunes on which agreement was high (top row) or low (bottom row) on the three structural modalities of interest (T1, FLAIR, T2)}
    \label{fig:ExEPVSLac}
\vskip -15pt
\end{figure}

The vascular network that supplies the brain changes with age, inducing alterations to surrounding tissue. Macroscopic changes can be observed on structural MR images and include white matter hyperintensities, lacunar infarcts, cerebral micro-haemorrhages and enlarged perivascular spaces (EPVS), among others. More specifically, perivascular spaces are thought to be used as a lymphatic pathway in a drainage mechanism, where entrapped fluid can extend this space, making it visible in MR images, often as linearly-shaped fluid-like structure. 
In clinical practice, their presence is classically assessed using visual scales on T2 MR images, described as elongated bright ellipsoids \cite{Potter2015}. The use of such visual scales requires extensive training and expertise, is prone to inter/intra rater variability, suffers from flooring/ceiling effects 
and is time-consuming for the operator. Some works have recently been proposed to automatically assess the EPVS burden \cite{Boespflug2018}\cite{Dubost2019} in clinical grade MR data, while others propose to segment EPVS at higher field (7T) \cite{Zhang2016}.
In contrast, lacunar infarcts, observed with a much lower frequency, are areas of dead tissue due to complete ischemia. Their shape signature is an ovoid object of $3-15mm$ of diameter, with a cerebrospinal fluid (CSF) -like intensity in the centre. Often, on T2 weighted Fluid attenuated inversion recovery (FLAIR) images, they are surrounded by a rim of hyperintensity.
In practice, even for trained radiologists, distinguishing between EPVS and lacunes can be very challenging. This results in double counting of uncertain objects \cite{DelC.ValdesHernandez2013}, and under-counting when objects branch from the same point (see Figure \ref{fig:ExEPVSLac}).

To account for the above-mentioned challenges, we propose to adapt the 2D RCNN model presented by He et al \cite{he2017mask} that allows for multiclass multi-instances simultaneous detection and segmentation to multirater 3D data, in the context of EPVS and lacune detection and size characterisation, with the perspective of a future expansion to more object classes (e.g.white matter hyperintensities) and their semantic segmentation. After a brief description of the 2D RCNN framework, we detail the challenges inherent to 3D data of such a framework in the capture of extremely small objects, and describe the introduction of multirater predictions.
\section{Methods}
\label{sec:methods}
\subsection{Two dimensional RCNN}
In the original RCNN framework, a backbone network is trained to extract generic features. This initial training is then complemented by two stages: a region proposal network and a final classification network applied to selected boxes whose shapes have been modified to fit a specified mask. 
In the 2D setting, the region proposal network is based on the classification as positive or negative of a series of predefined boxes created based on anchors, regularly spaced on the 2D grid with different ratios of height and width. All selected grid are then resampled (pooled) to a user-specified shape and fed to the final segmentation classification branch of the framework.

\subsection{Challenges and strategies for a multirater 3D extension}
The main challenges related to the extension of the successful RCNN framework to 3D data lay in the memory and data requirements, as well as an extreme class imbalance. In terms of memory, the generation of grid anchors become notably prohibitive in 3D. Additionally, when dealing with ESOs, any interpolation induced by the region pooling may obscure relevant features and render the segmentation meaningless. 
In order to account for these challenges, the following strategies were adopted at the different stages of the framework:
\paragraph{Backbone network}
The 3D HighResNet proposed by Li et al. \cite{Li} was used as backbone network to extract features. This architecture has a large contextual field of view at reduced parameter cost. This network uses three levels of residual convolutional networks with dilated convolutions with increasing dilation factor, each level consisting of three dilated convolutions with fixed dilation factor alternating with batch normalisation and ReLu activation. In the presented setting, the network was applied to regress a distance map with a root mean square error loss. The distance map is calculated from each given element's segmentation. 
 
 \paragraph{Region Proposal Network (RPN)}
 In order to alleviate the memory burden of having to explicitly describe anchors and associated boxes, the RPN, consisting of one classification and one regression branch, was applied in a convolutional fashion to every voxel. The features extracted at the backbone level were fed into a small convolutional network with a single common $3^3$ kernel, followed by either a classification layer or a regression layer. The classification layer establishes if the centre of the patch is likely to be the centre of mass of the target object, while the regression part outputs four values: three values representing the distance to the closest object centre of mass, and the fourth representing the scale of the targeted object. Classification and regression were learnt from 300 samples on the patch, with a 50/50 balance between positive and negative samples. To avoid any impact on the regression branch, negative samples did not bear any weight on the regression loss. A cross-entropy loss was used for the classification branch while a smooth distance loss was applied on the regression branch for the estimation of the distance to the closest element centre of mass. Denoting $r_{n}$ the absolute error between predicted value and ground truth for a given sample $n$, the smooth distance loss $DL$ is expressed as:
 \[DL=\dfrac{1}{N}\sum_{n=1}^{N}f(r_n) \text{ where } f(r_n)=\left\{\begin{array}{cc} 0.5 r_n^{2} &\text{ if }r_n < 0.5\\
 (r_n - 0.125)^{2} -2 & \text{ if } r_n>2.125 \\
 r_n - 0.125 & \text{ otherwise}\\
 \end{array}\right.\]
 
 \paragraph{Refinement/Classification Network (RCN)}
 From the location of proposed ESO centres-of-mass, boxes were associated with ground-truth objects, and extracted masks are directly fed so as to classify the boxes and adjust the regression of the centre of mass. 

 The branch jointly classifying the element and regressing centre of mass and object scale consisted of a convolutional layer of kernel size 7, followed by a fully connected layer. After average pooling, classification and bounding box regression were established. For the regression branch, the target prediction was the residual between the RPN prediction and the ground truth for the three location elements, and a scale correction factor for the size. A similar smooth distance loss was applied as a cost function.
 In contrast to the original RCNN framework, selected boxes were neither resized nor pooled to a predefined shape. This is in order to avoid interpolation that would be detrimental, given that many of the targeted elements are one voxel wide.
 \begin{figure}[b!]
     \centering
     \includegraphics[trim=1cm 3cm 0cm 3cm, clip=true, width=0.9\textwidth]{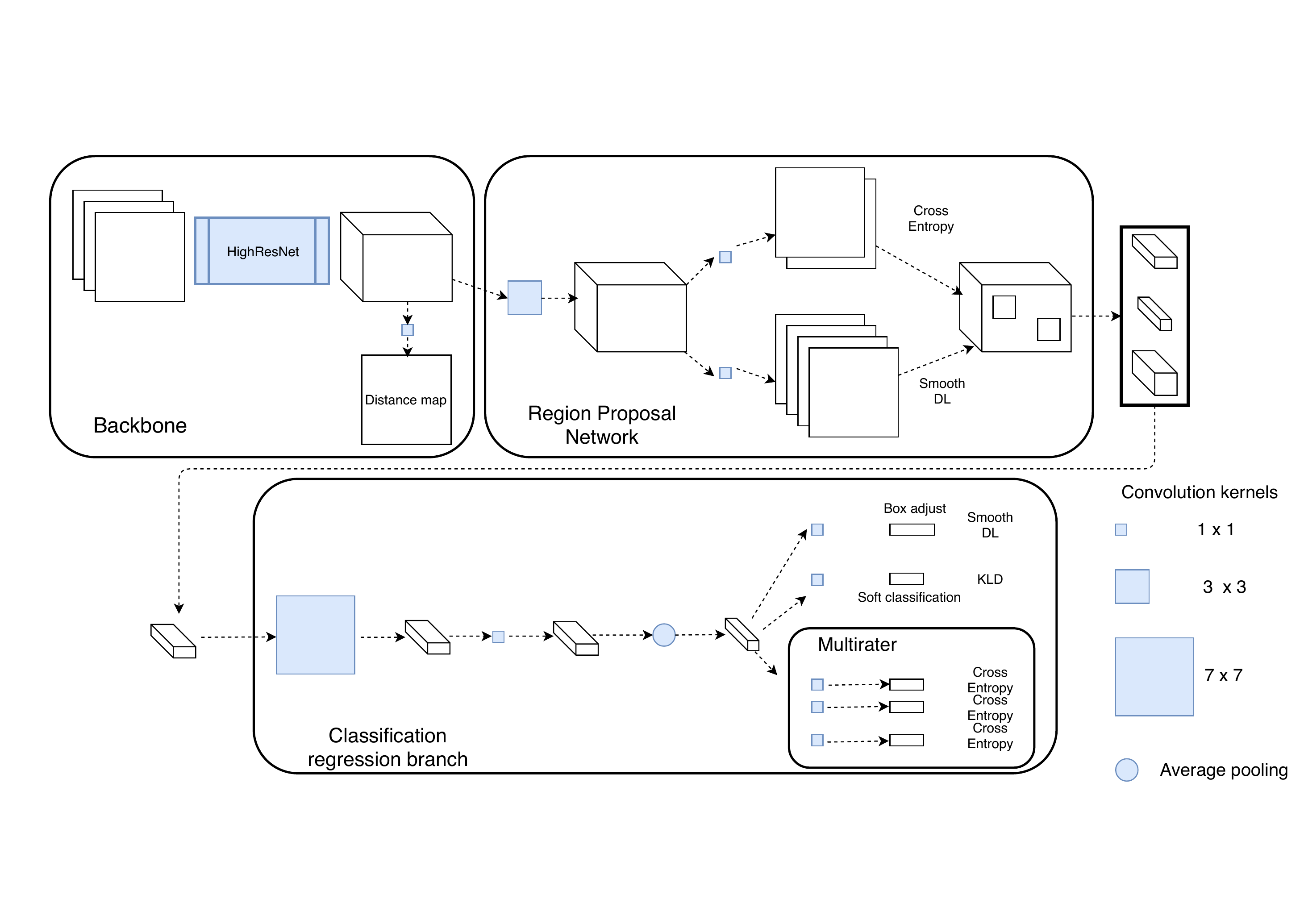}
     \caption{Architecture of the 3D multirater RCNN for extremely small objects.}
     \label{fig:FrameworkArchitecture}
 \vskip -10pt
 \end{figure}
 \paragraph{Multirater encoding}
 For each of the manually-segmented elements, the raters were asked to attribute one of the following class: 1)Nothing; 2) Lacune; 3) EPVS; 4) Undecided between lacune and EPVS; 
 Instead of a crisp classification, a soft probability label was obtained as the average of the multiple raters involved in the classification and used as target. For each rater, a fully connected layer was added in order to directly infer the classification of each individual. The architecture framework is displayed in Figure \ref{fig:FrameworkArchitecture}.

\subsection{Implementation}
\paragraph{Sampling and data normalisation}
The existence of two types of imbalance (foreground vs background, and between EPVS vs lacunes) required a purpose-specific sampling scheme. A probabilistic weight sampling was adopted as suggested by Ronneberger et al \cite{ronneberger2015u} to extract patches of size $64^3$ over the images. For this purpose, the inverse of the distance maps from segmented EPVS and lacunes were smoothed and linearly combined using a ratio of 1/100 reflecting the relative frequency of occurrence of these two classes. These maps were clipped to a minimum of $10^{-5}$ to reflect the overall background/foreground ratio. 
All input data was bias field corrected, skull stripped, and then z-scored to the white matter region statistics. 
\paragraph{Training scheduling and loss functions}
The framework was implemented within NiftyNet~\cite{gibson2018niftynet} (\url{niftynet.io}) and will be merged into the main codebase at the time of publication.
The network was trained progressively per stage to mitigate training stability issues. Sections where classification and regression were combined (RPN and RCN) were trained in two steps: the first one consisted of the classification training with a sigmoid applied to the regression loss, and the second step was the sum of the two losses. Each of the steps was trained for 1000 iterations with learning rate of 0.0001. In order to account for scale differences observed across combined loss functions, notably between classification and positioning regression losses, empirical weights 
were chosen and progressively modified throughout the training of the network in order to always ensure a balance between classification accuracy and box positioning.
\paragraph{Inference}
At inference, a similar patch size was used as for the training step in order to expect a similar number of proposals (limited to 300). Having obtained the distance regression map and the score map, the 300 proposed centres of mass were extracted based on the skeleton maxima of the smoothed regressed distance map (p score map $>$0.25). Centres of mass closer than 2mm were pruned as a form of non-maximum suppression. 

\section{Data and experiments}
\subsection{Data}
\label{sec:data}
\begin{figure}[b!]
    \centering
    \begin{tabular}{cc}
    \includegraphics[trim={0cm 0cm 0cm 0cm}, clip=true, width=0.30\textwidth]{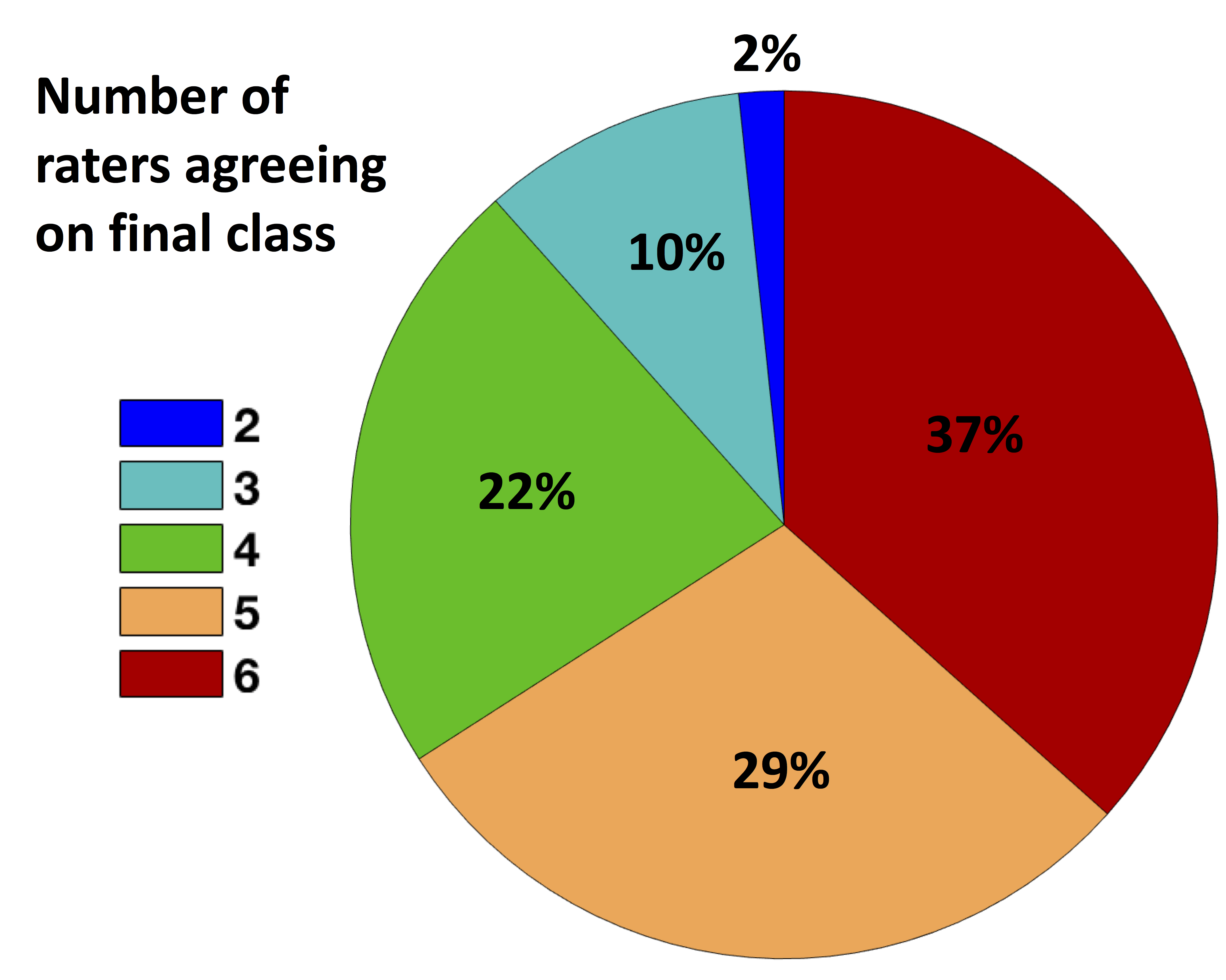} & \includegraphics[trim={2cm 0.5cm 2cm 1cm}, clip=true, width=0.60\textwidth]{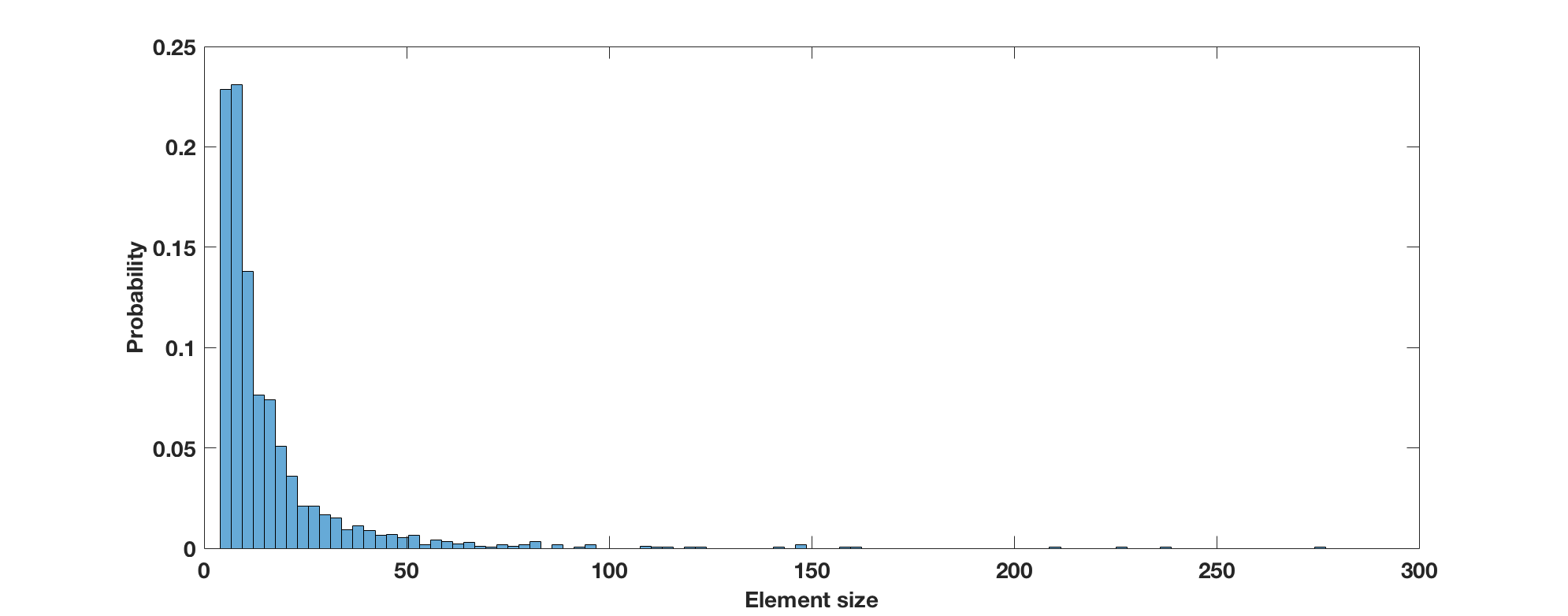}
    \end{tabular}
    \caption{Repartition of agreement between raters responsible for the final crisp classification (left) and distribution of the size of the targeted elements (right).}
    \label{fig:datainfo}
\end{figure}
16 subjects were selected out of a longitudinal tri-ethnic cohort of elderly subjects aiming at investigating the relationship between cardiovascular risk factors and brain health \cite{Tillin2012SABRECohort}. At the third wave of investigation, subjects of this cohort underwent an MR session including the acquisition of 3D isotropic T1 weighted, T2 weighted and T2-weighted FLAIR images \cite{Sudre2018}. The 16 subjects were chosen for their elevated vascular burden visually assessed by a trained radiologist. EPVS and lacunes were manually segmented using the three available structural MR sequences using ITKSnap \cite{py06nimg}. Segmentations were done in a multi-view manner to ensure geometrical consistency, with all images aligned to the T1 sequence as a geometrical reference. Segmentation masks were then automatically corrected and voxels with inappropriate signal identity signature were removed. Individual EPVS and lacunes were further classified at the level of connected components by six operators with a varied range of expertise using an in house dedicated viewer. Only elements with a volume of more than 5 voxels were further used for the training of the network. Out of the initial 4147 considered elements, 2442 were used as gold standards for training. The volumes of segmented elements ranged thus from 5 to 350, with 48.8\% with a size below 10 voxels. 
Perfect agreement among raters was reached only in 36.6\% of the cases, and only 2.8\% of the elements were ultimately classified as lacunes. Figure \ref{fig:datainfo} presents an histogram of element size, and a pie chart representing the proportion or rater agreement. The poor inter-rater classification agreement hints at the complexity of the task. Uncertainty over the segmentation would have to be evaluated over multiple raters before envisioning moving the proposed object RCNN detection model to a full Mask-RCNN, also performing segmentation. Due to the lack of more training data, 14 of the subjects were used for training and 2 were hold-out for testing.

\subsection{Experiments}
\label{sec:experiment}
In order to compare the performance of a standard segmentation approach to the proposed multiclass detection framework, we trained semantic segmentation models with multiple combinations of architectures, loss functions (e.g Generalised Dice Loss), learning rates (from $10^{-6}$ to $10^{-3}$) and regularisation. Parameter choice was similar to the one used for the backbone network, with ranges that have been shown to perform well on unbalanced data. Unfortunately, no network was able to segment any foreground class.  

We present hereafter the results obtained at the different stages of the model in terms of distance regression, score map, RPN and multirater classification. 

\section{Results}
\label{sec:results}
\begin{figure}[b!]
\vskip -10pt
    \centering
    \setlength{\tabcolsep}{0.1em}
    \begin{tabular}{cccccc}
    T1 & FLAIR & T2 &  Ground truth & Distance map & Score map\\
    \includegraphics[trim=6.5cm 2cm 13.5cm 2cm, clip=true, width=0.16\textwidth]{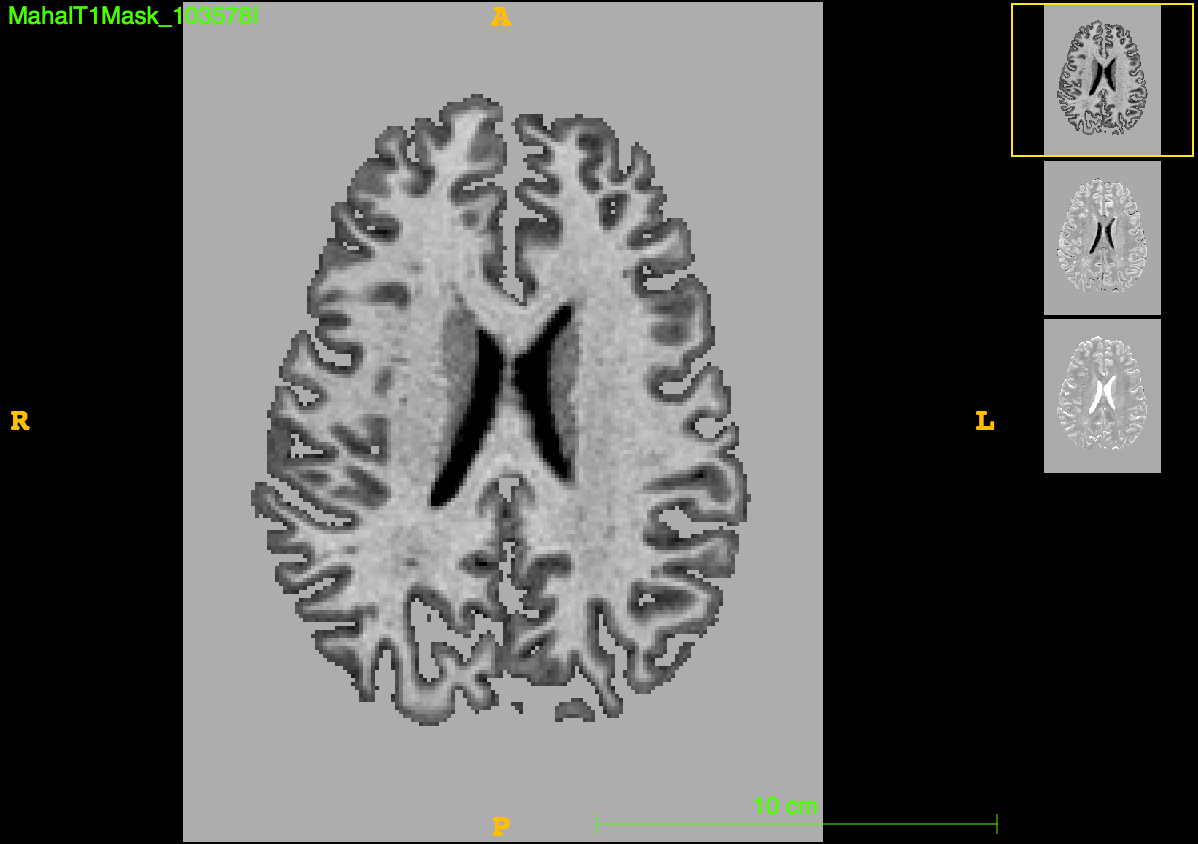}& \includegraphics[trim=6.5cm 2cm 13.5cm 2cm, clip=true, width=0.16\textwidth]{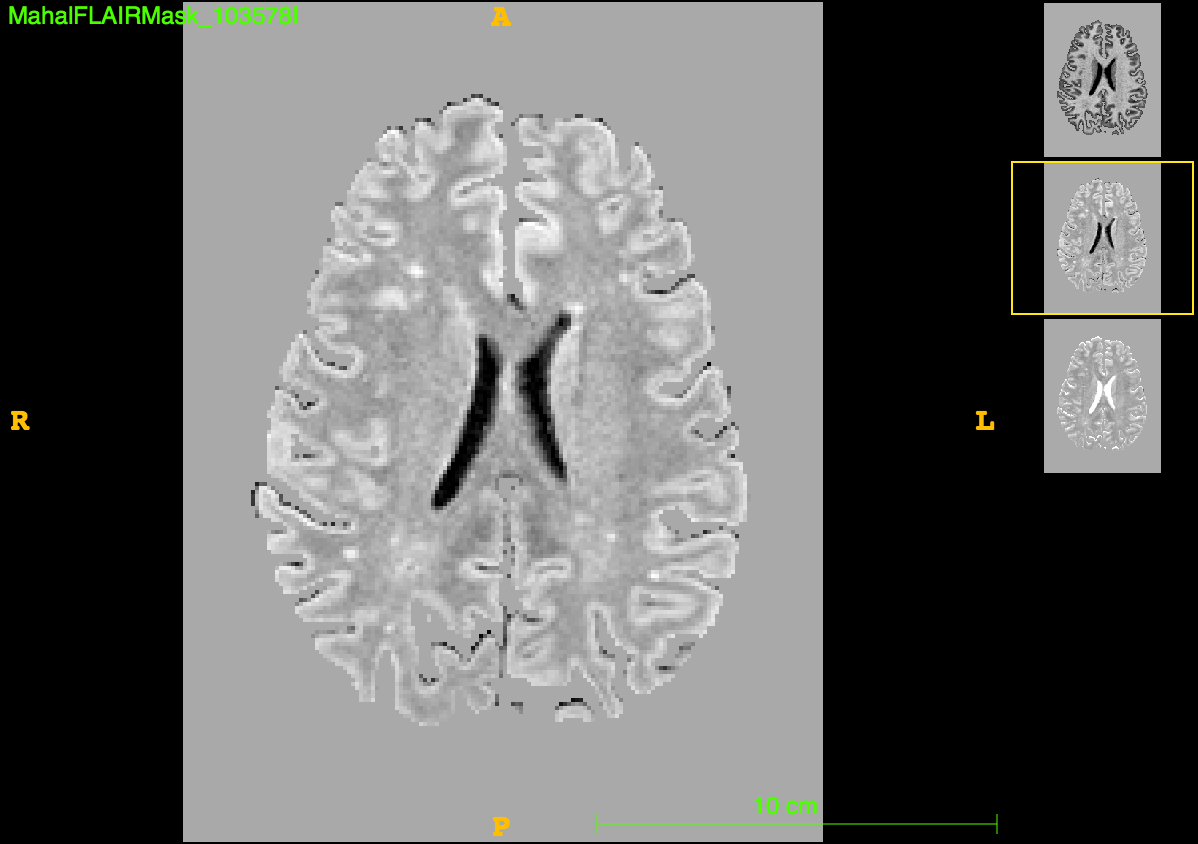}& \includegraphics[trim=6.5cm 2cm 13.5cm 2cm, clip=true, width=0.16\textwidth]{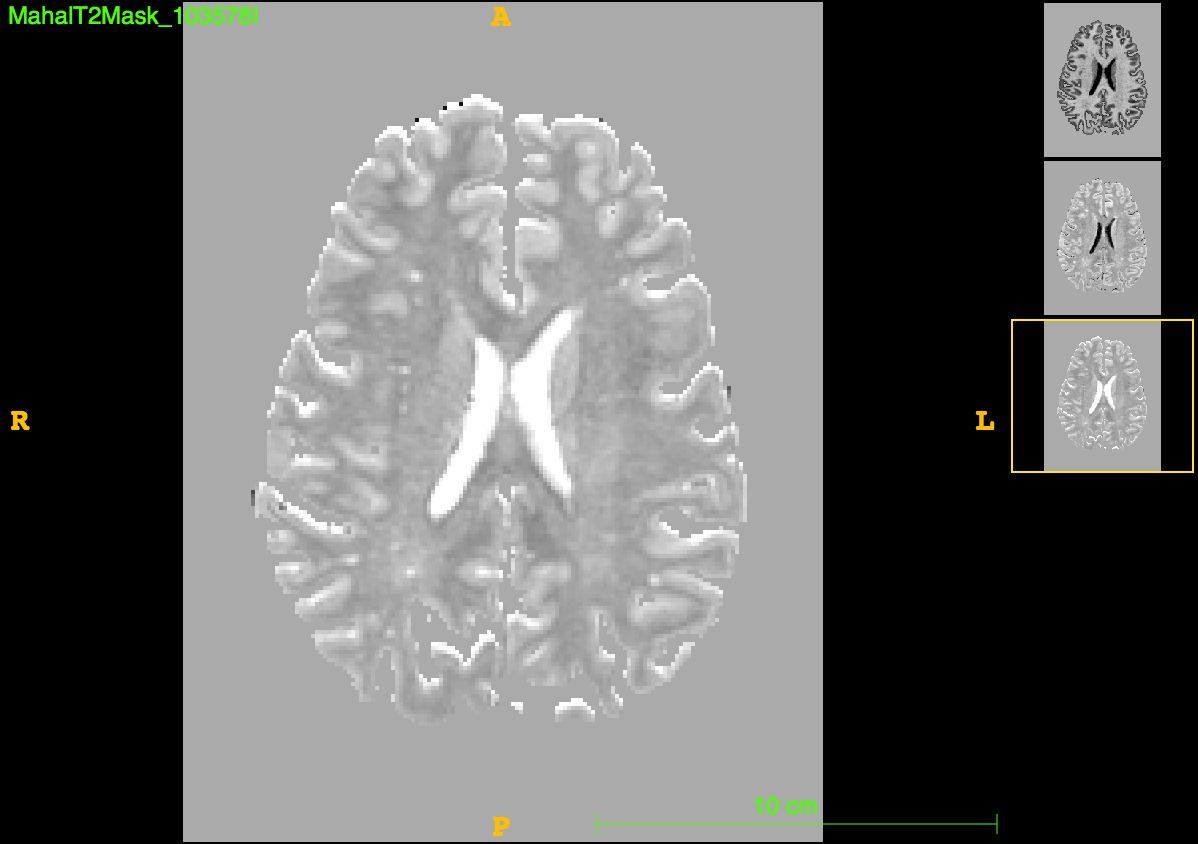}&
    \includegraphics[trim=6.5cm 2cm 13.5cm 2cm, clip=true, width=0.16\textwidth]{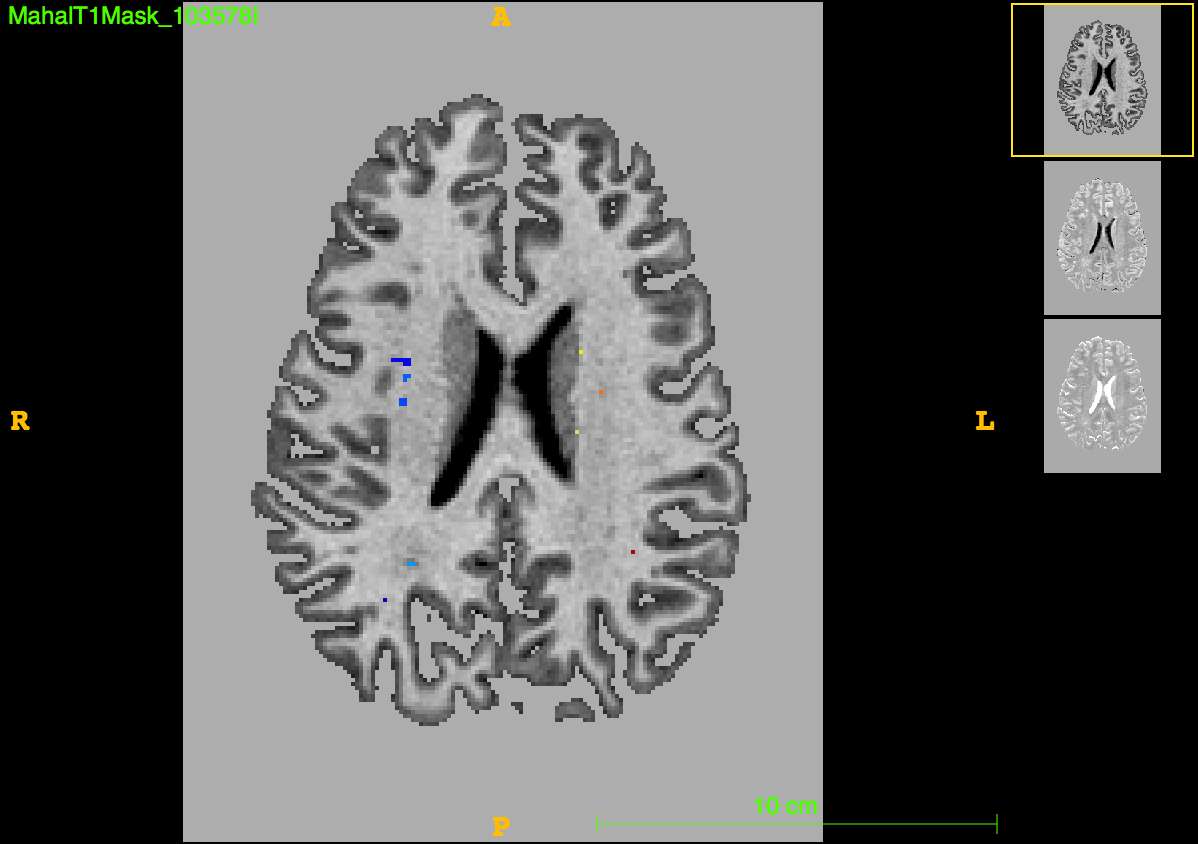}& \includegraphics[trim=6.5cm 2cm 13.5cm 2cm, clip=true, width=0.16\textwidth]{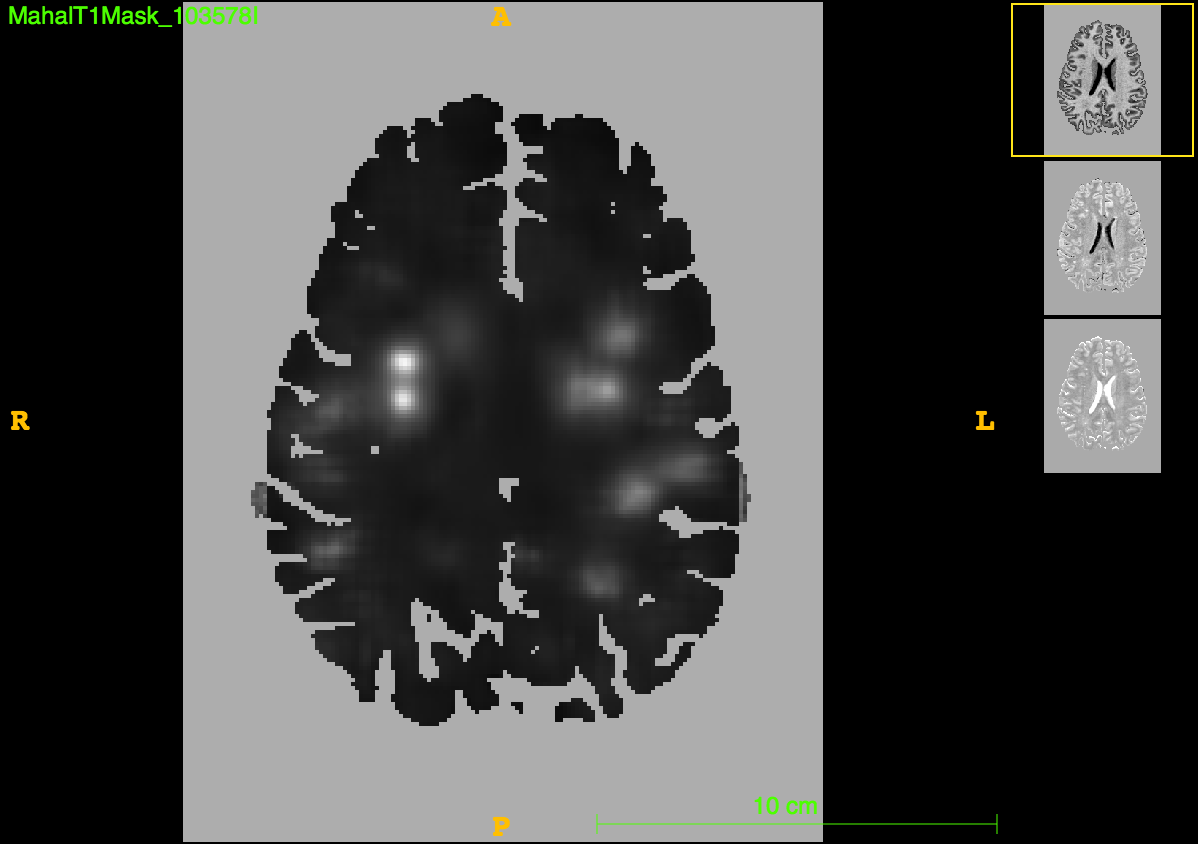}& \includegraphics[trim=6.5cm 2cm 13.5cm 2cm, clip=true, width=0.16\textwidth]{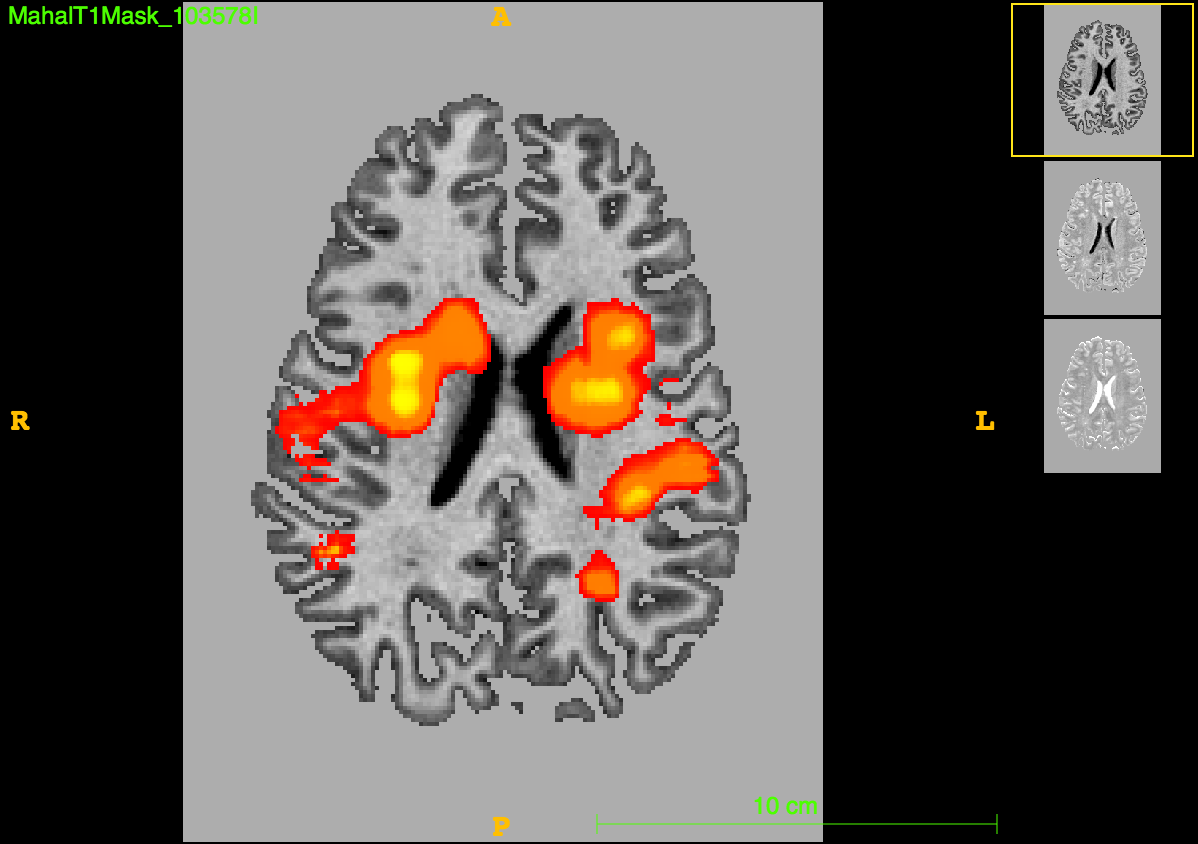}\\
    \includegraphics[trim=6.5cm 2cm 13.5cm 2cm, clip=true, width=0.16\textwidth]{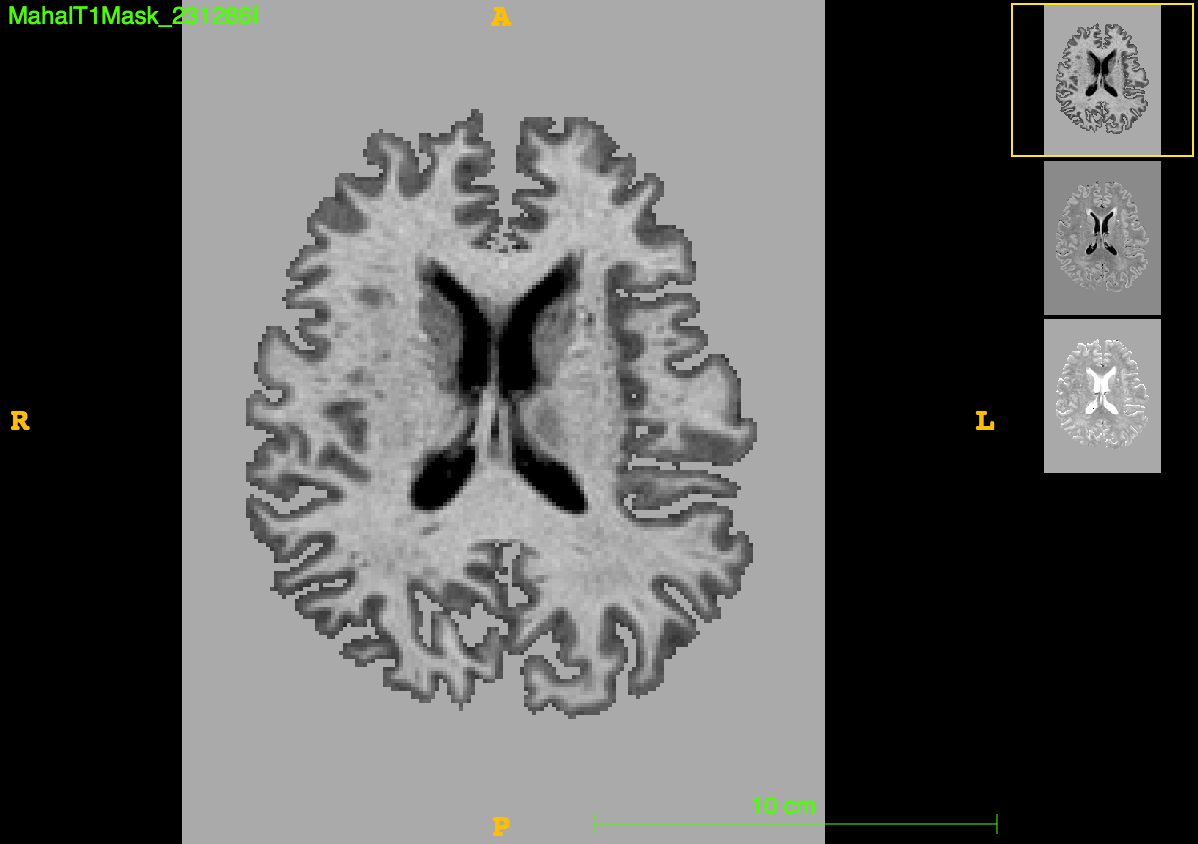}& \includegraphics[trim=6.5cm 2cm 13.5cm 2cm, clip=true, width=0.16\textwidth]{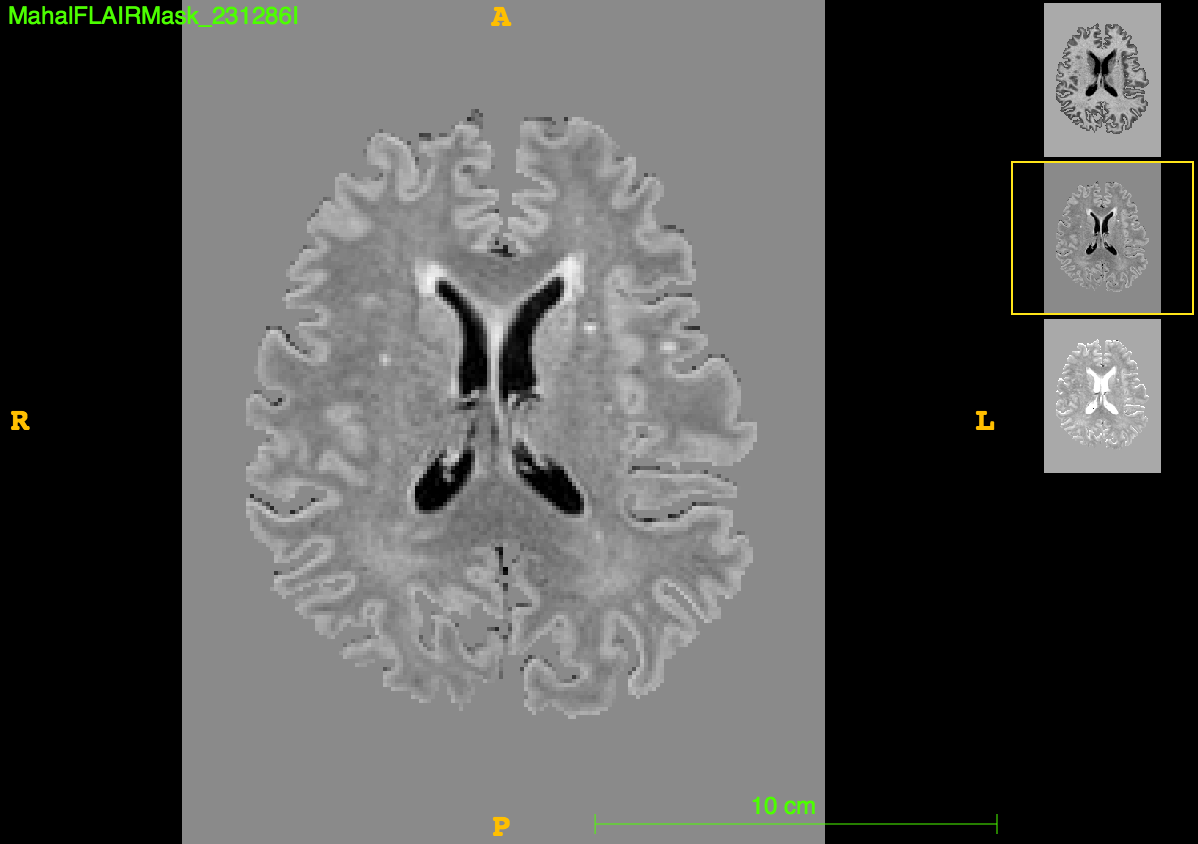}& \includegraphics[trim=6.5cm 2cm 13.5cm 2cm, clip=true, width=0.16\textwidth]{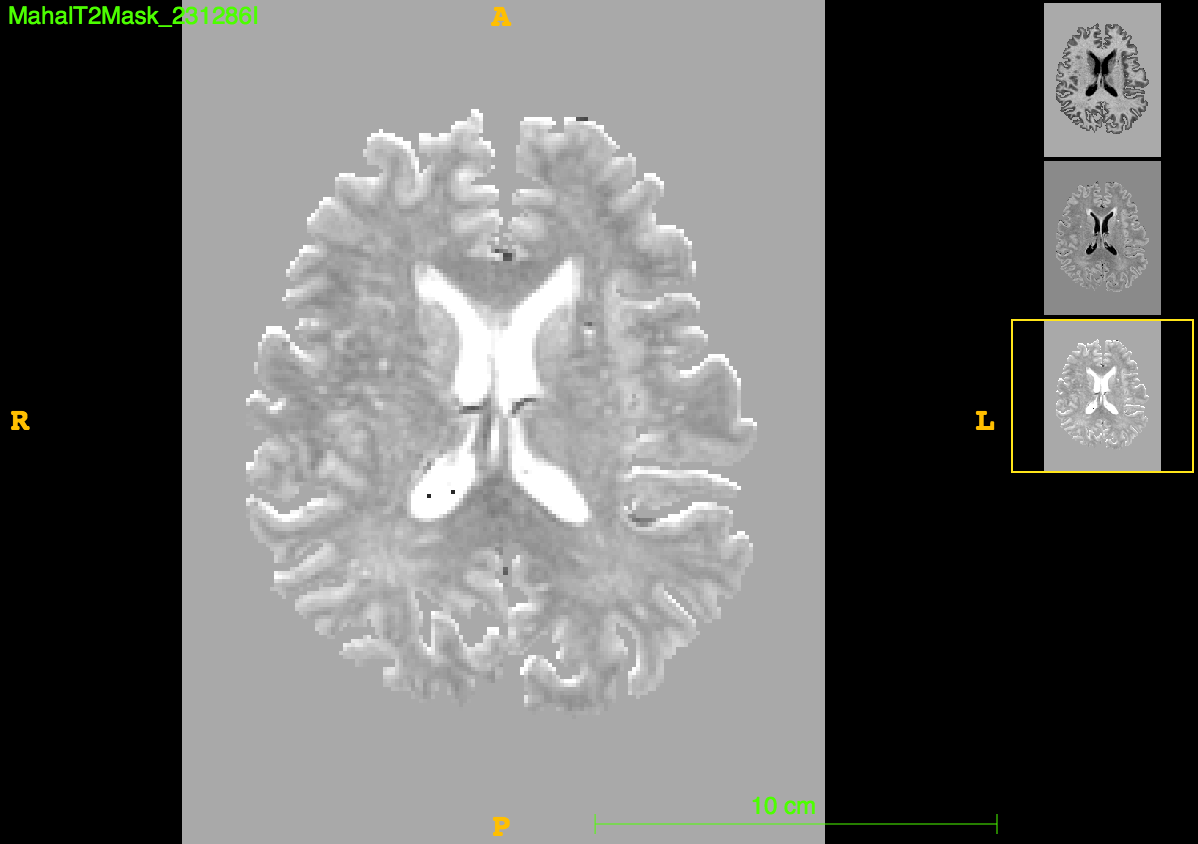}&
    \includegraphics[trim=6.5cm 2cm 13.5cm 2cm, clip=true, width=0.16\textwidth]{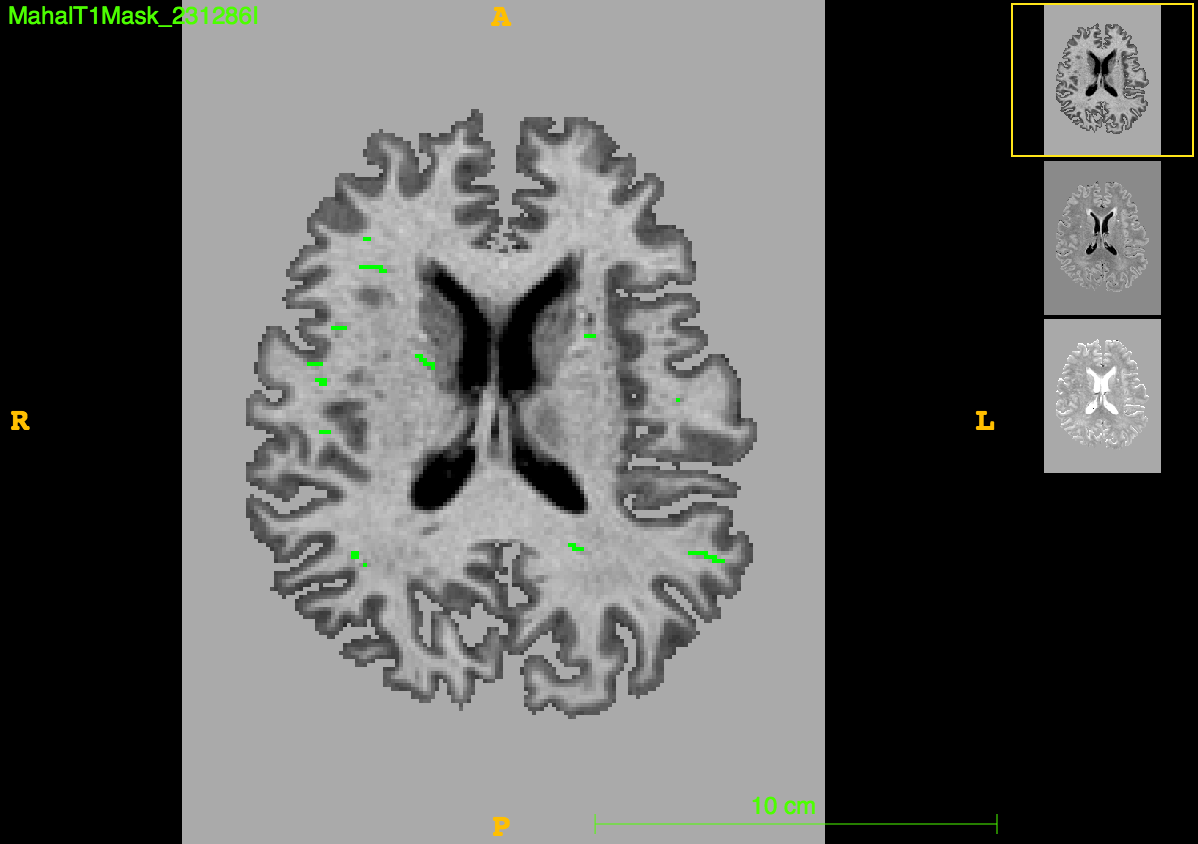}& \includegraphics[trim=6.5cm 2cm 13.5cm 2cm, clip=true, width=0.16\textwidth]{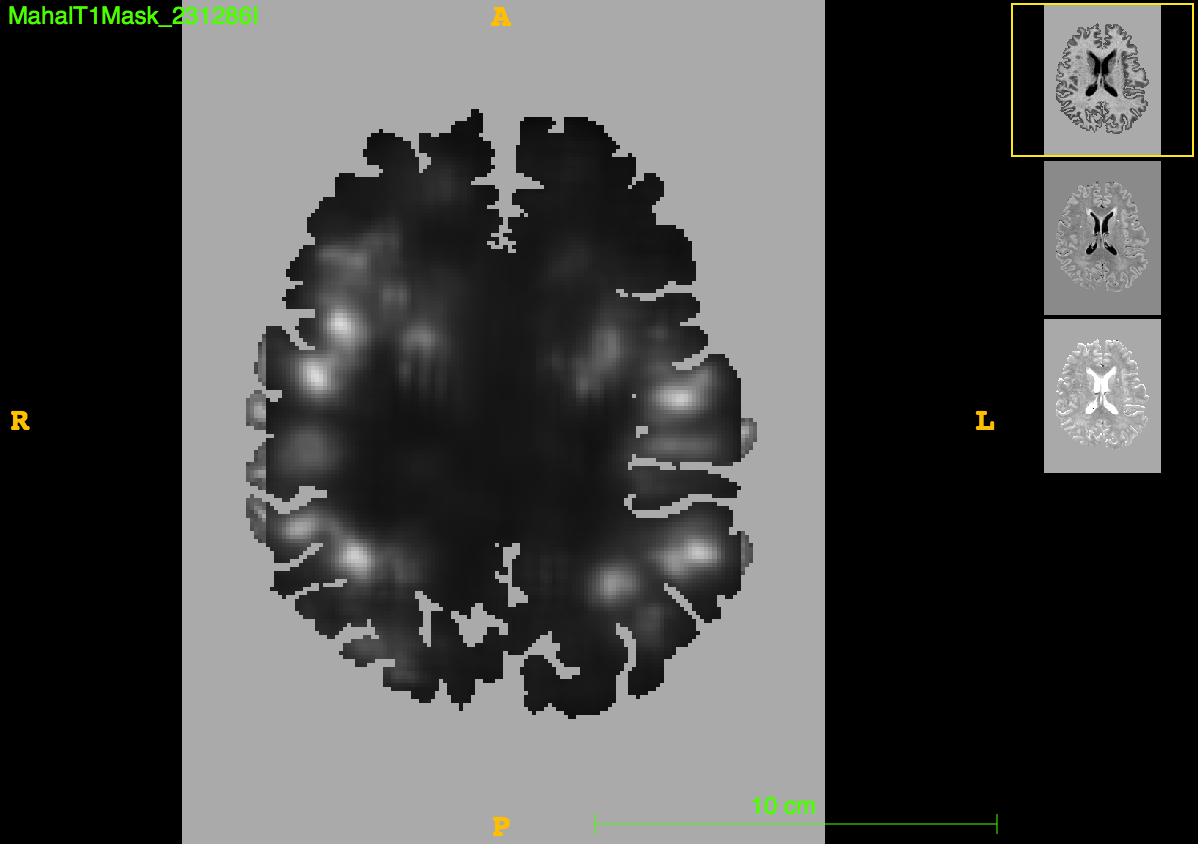}& \includegraphics[trim=6.5cm 2cm 13.5cm 2cm, clip=true, width=0.16\textwidth]{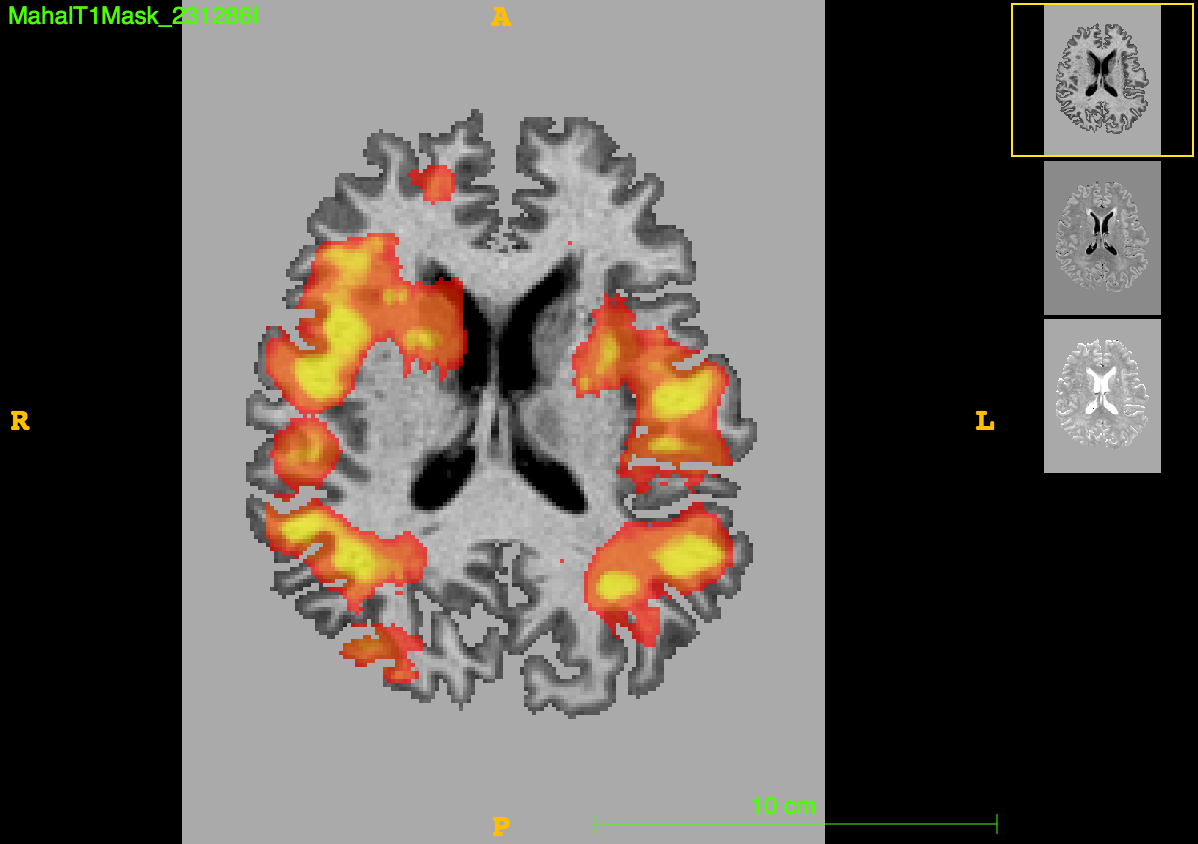}\\
    \end{tabular}
    \caption{Two holdout cases with the three input channels (T1, FLAIR, T2), gold standard segmentation, inferred distance maps and score map }
    \label{fig:DistanceScore}
    \vskip -20pt
\end{figure}
\begin{figure}[t!]
    \centering
    \begin{tabular}{ccc}
    T1 & Score map & Extracted boxes \\ 
    \includegraphics[trim=7cm 3cm 2cm 6cm, clip=true, width=0.3\textwidth]{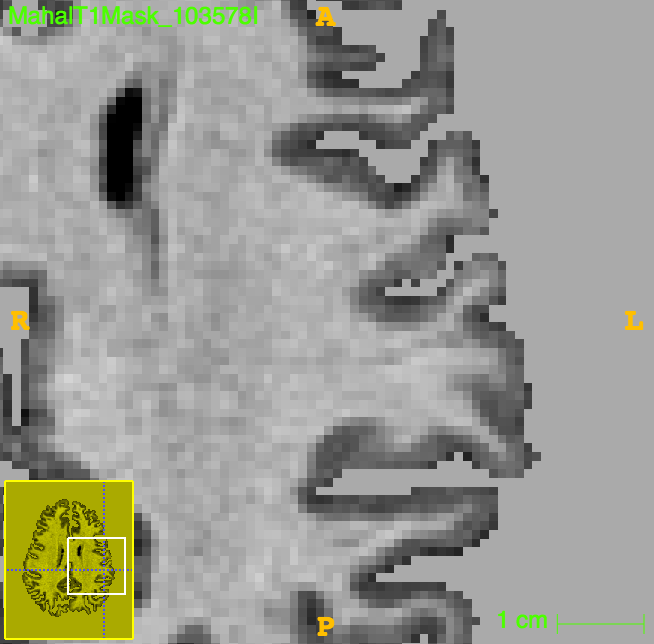} & \includegraphics[trim=7cm 3cm 2cm 6cm, clip=true, width=0.3\textwidth]{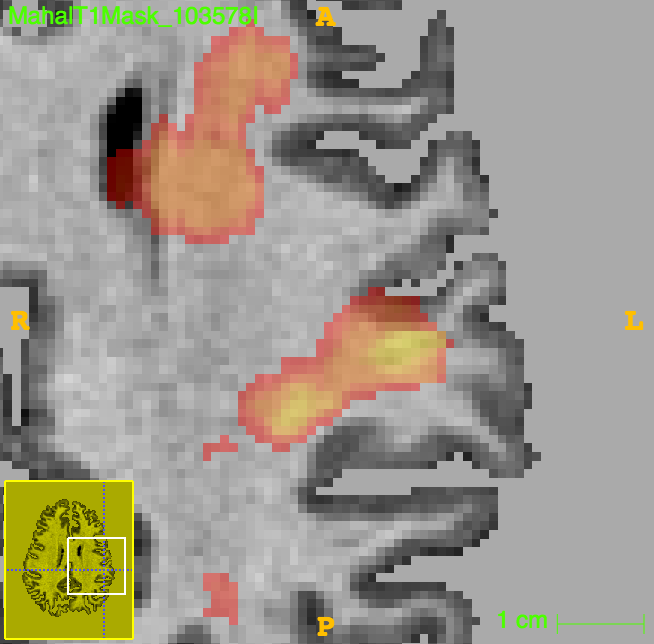}& \includegraphics[trim=7cm 3cm 2cm 6cm, clip=true, width=0.3\textwidth]{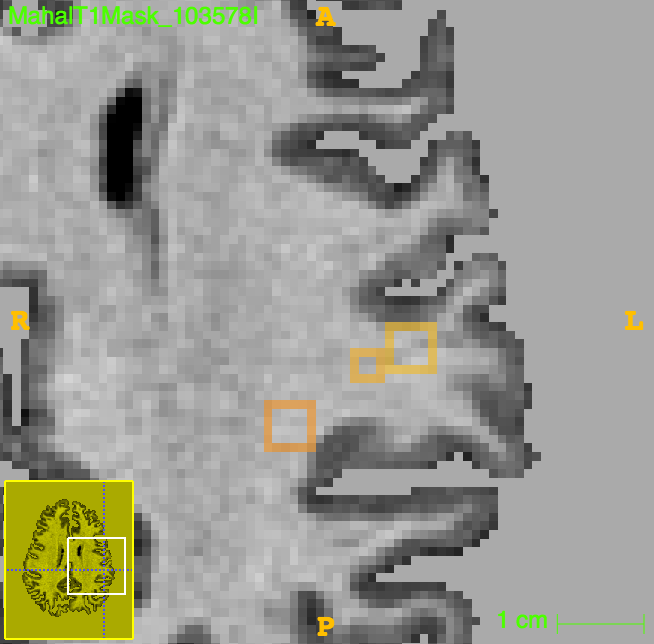}\\
    \end{tabular}
    \caption{ESOs rightly detected by the network but missed during manual labelling. From left to right, T1, predicted score map and predicted boxes.}
    \label{fig:ErrorSegmentation}
    \vskip -15pt
\end{figure}

Each step of the framework was assessed on the two held out test subjects using the same metrics as the loss functions.
Figure \ref{fig:DistanceScore} presents the input data for the three modalities along with the ground truth segmentation, the regressed distance map and the inferred score map.
Interestingly, some elements not present in the gold standard segmentation but detected as per the score map were a posteriori considered as valid enlarged perivascular spaces as can be seen on Figure \ref{fig:ErrorSegmentation}.

\begin{figure}[b!]
    \centering
    \begin{tabular}{cc}
    \includegraphics[width=0.4\textwidth]{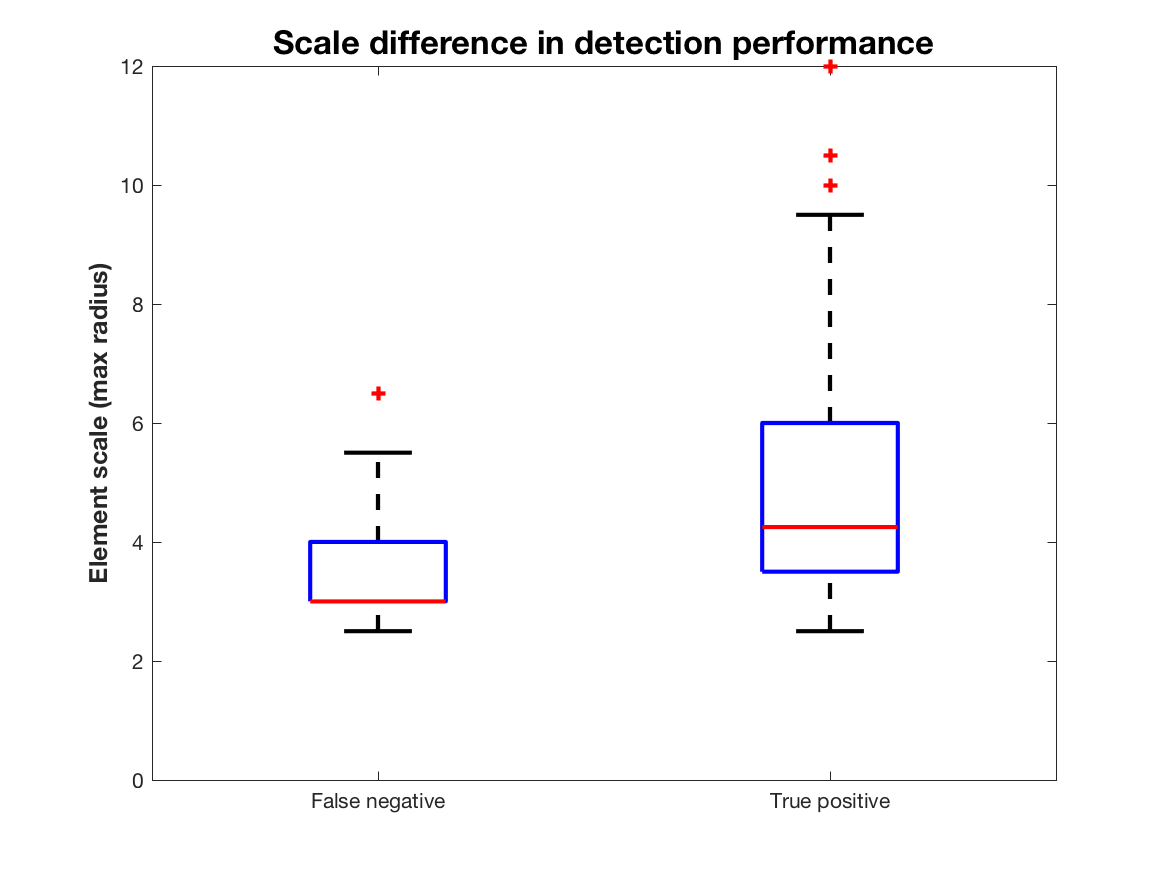} & \includegraphics[width=0.4\textwidth]{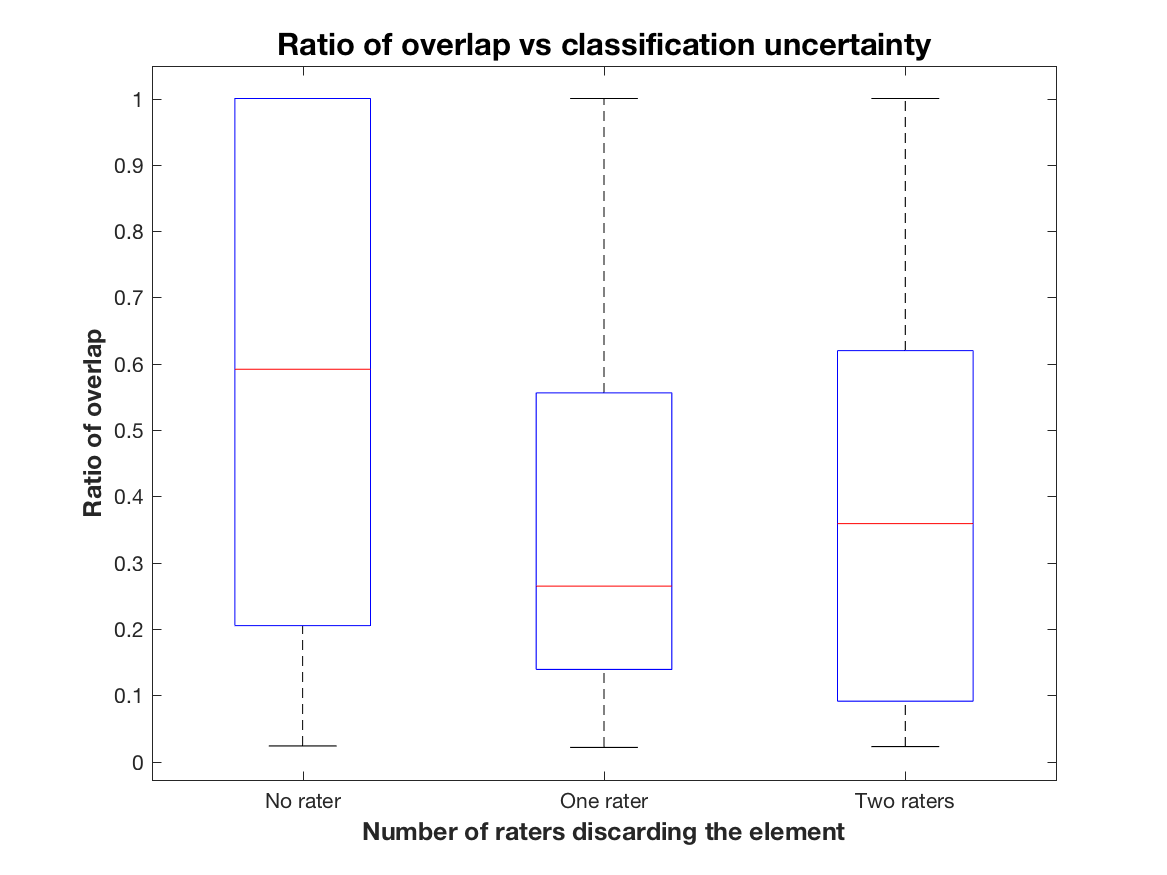}
    \end{tabular}
    \caption{Left: Gold standard scale of ESO versus detection (False negative/True positive). Right: Relationship between multi-rater disagreement and box overlap performance. Note that overlap ratio is higher for more certain objects.}
    \label{fig:boxplot_uncertainty}
    \vskip -15pt
\end{figure}
Given the limitations of the available gold standard in terms of inter-rater element classification, and potential missing objects, the validation focused on the sensitivity of the trained model and the relationship of the results with the multi-rater uncertainty.
A sensitivity of 72.7\% was observed across the two test subjects with a significant difference in element size between false negatives and true positives (Wilcoxon ranksum test p$<$0.00001). Investigating the relationship between the ratio of overlap between best matching detected box and ground truth proposal, a significant association between agreement of raters and overlap was observed (p=0.002) with a median overlap of 59\% when all raters agreed and an overlap of 30\% for the more uncertain cases (at least one rater considering the element not to be relevant). Note that overlap is measured on the predicted box, which can vary widely in its size. Figure \ref{fig:boxplot_uncertainty} presents boxplots of relationship between ESO scale and detection (left), and overlap ratio with rater uncertainty (right).

\begin{figure}[t!]
\vskip -15pt
    \centering
    \begin{tabular}{ccccc}
    & Nothing & Lacune & EPVS & Undecided\\
    GT&\raisebox{-.5\height}{\includegraphics[trim=3.5cm 2cm 3.5cm 2cm, clip=true, width=0.15\textwidth]{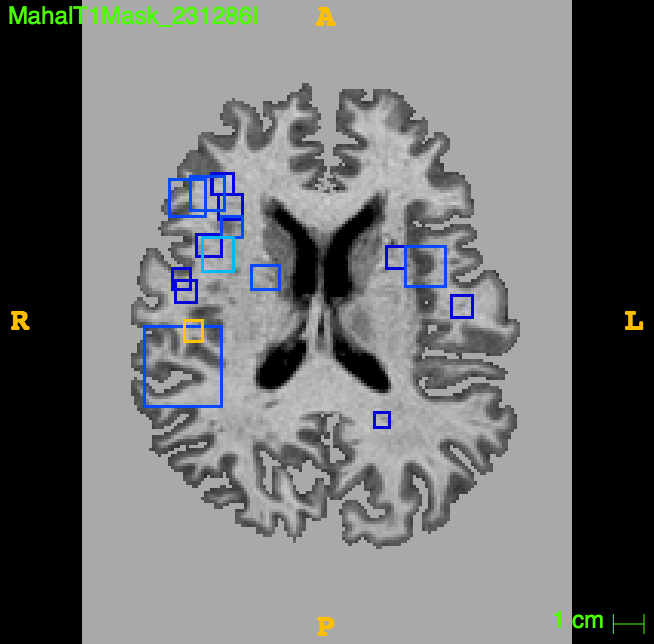}} & \raisebox{-.5\height}{\includegraphics[trim=3.5cm 2cm 3.5cm 2cm, clip=true, width=0.15\textwidth]{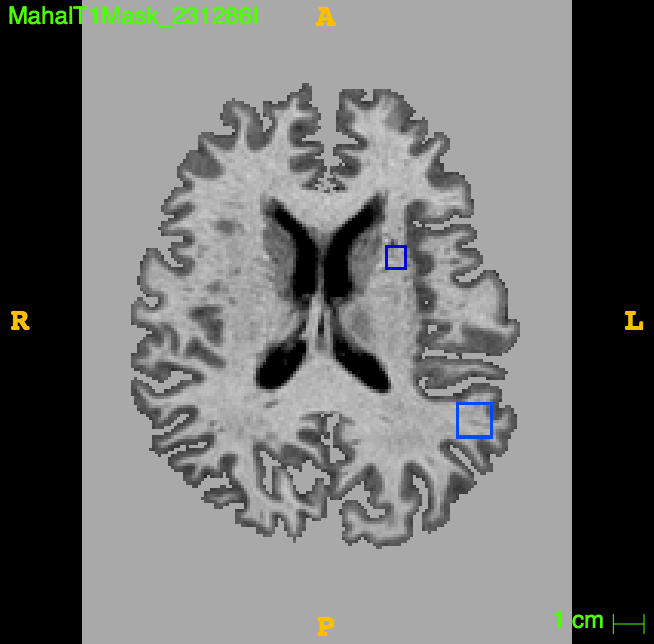}} & \raisebox{-.5\height}{\includegraphics[trim=3.5cm 2cm 3.5cm 2cm, clip=true, width=0.15\textwidth]{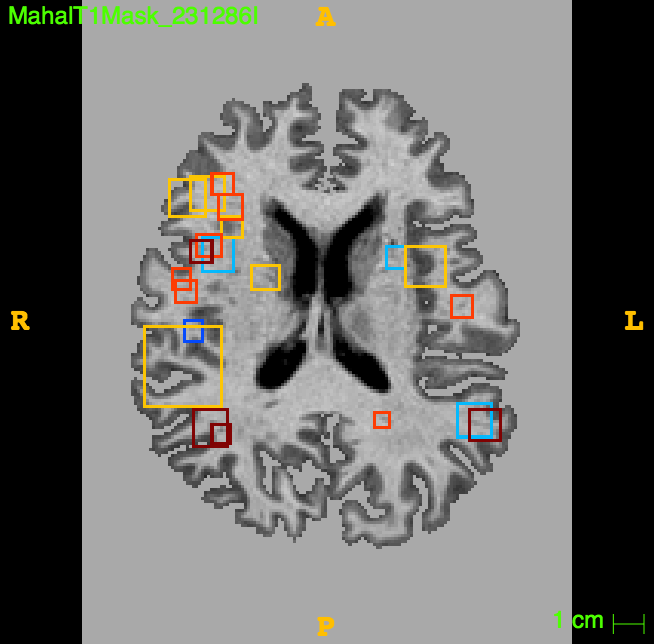}} & \raisebox{-.5\height}{\includegraphics[trim=3.5cm 2cm 3.5cm 2cm, clip=true, width=0.15\textwidth]{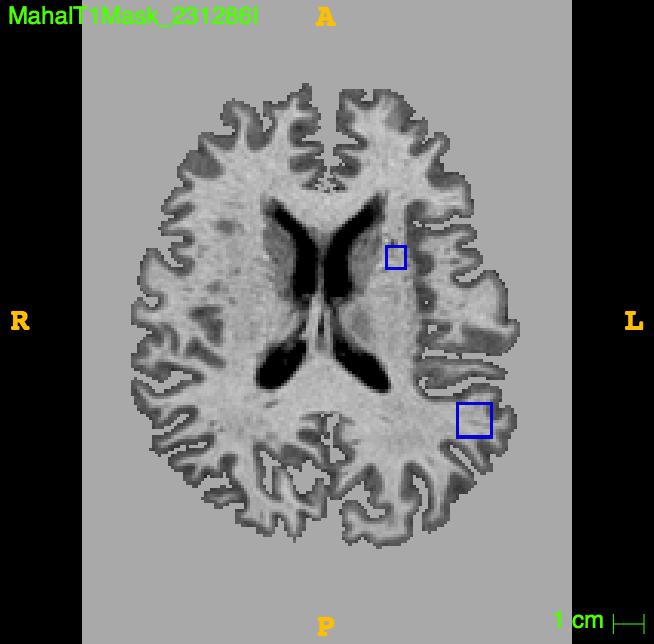}}\\
       Predicted& \raisebox{-.5\height}{\includegraphics[trim=3.5cm 2cm 3.5cm 2cm, clip=true, width=0.15\textwidth]{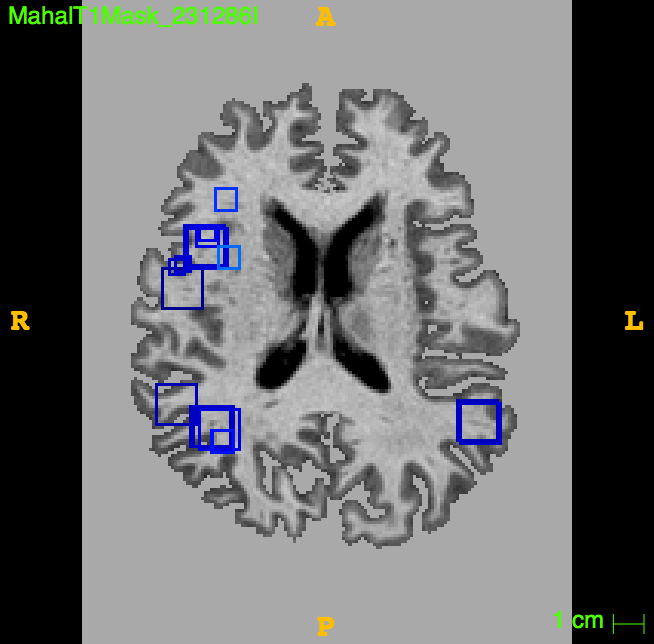}} & \raisebox{-.5\height}{\includegraphics[trim=3.5cm 2cm 3.5cm 2cm, clip=true, width=0.15\textwidth]{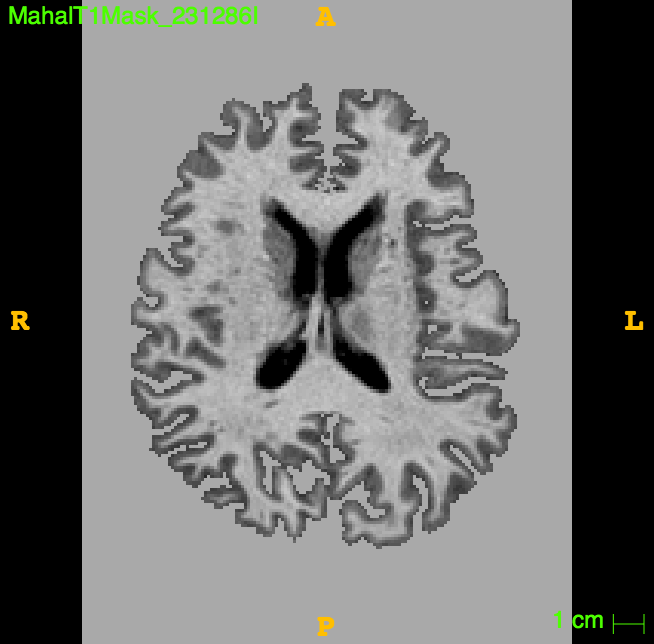}} & \raisebox{-.5\height}{\includegraphics[trim=3.5cm 2cm 3.5cm 2cm, clip=true, width=0.15\textwidth]{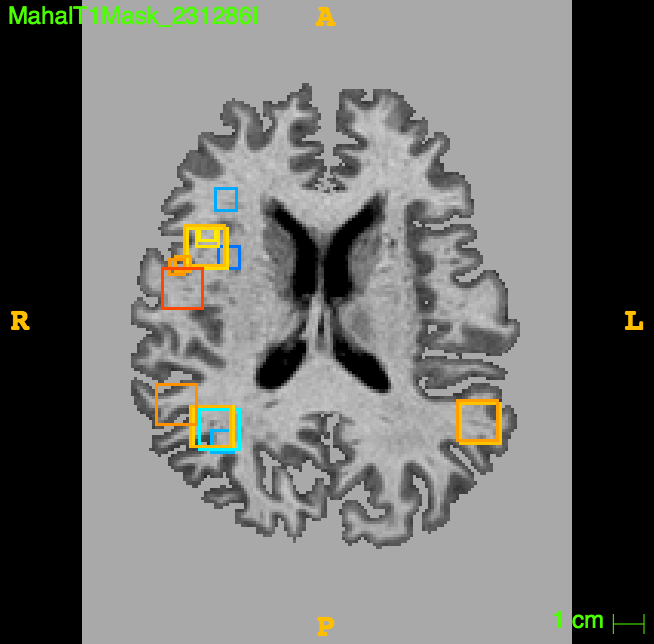}} & \raisebox{-.5\height}{\includegraphics[trim=3.5cm 2cm 3.5cm 2cm, clip=true, width=0.15\textwidth]{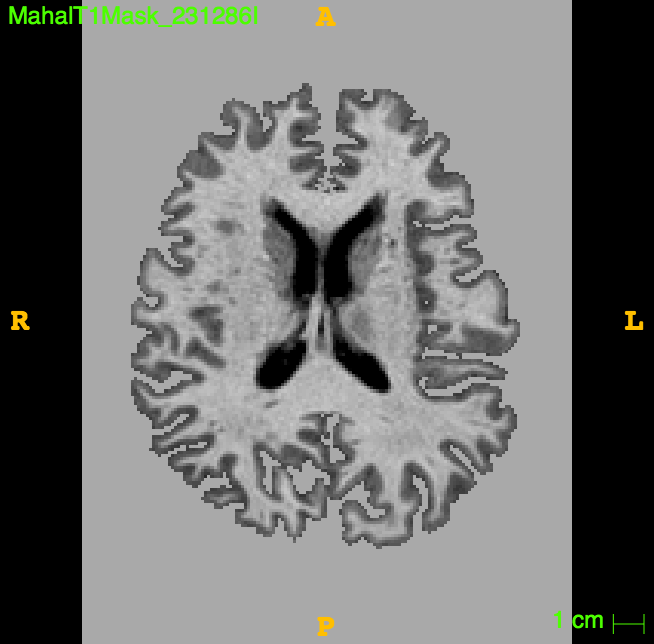}} \\
        \end{tabular}
    \caption{Ground truth and predicted probabilistic boxes for the different classes. All blue boxes correspond to very uncertain classification (p$<$0.5), and should be disregarded. Yellow to red boxes represent probabilities ranging from 0.5 to 1, and represent confident ESOs. }
    \label{fig:box_figure}
\end{figure}

Figure \ref{fig:box_figure} presents the ground truth and matching predicted boxes where the color reflects the probability of belonging to each of the classes (nothing - lacune - EPVS - undecided).

\section{Discussion and conclusion}
\label{sec:discussion}
In this work we proposed a 3D deep learning model for the detection and characterisation of extremely small objects incorporating multi-rater labels and agreement. In this context, two types of extreme class imbalance were found, with a very low ratio of foreground to background, as well as a strong imbalance between the estimated classes where the prevalence of enlarged perivascular spaces being much higher than the number of lacunes.

The different steps of the framework were evaluated, showing a good sensitivity of the region proposal network. Specificity was not ideal, probably limited by the missing annotation of individual branching elements (currently considered as a single ESO). 
Future work will use the multi-rater gold standard to better guide network updates by penalising classification errors made on definite classifications more strongly. Additionally, the segmentation, currently only used to obtain the original distance map, could enrich the model by defining a soft labelling at the edges and/or obtaining additional manual ratings. Furthermore, it must be noted that the training accuracy heavily depends on the quality of the initial co-registration of the different modalities, as one voxel of shift may lead to an aberrant intensity signature.
At this stage, proposal boxes are cuboid, since a single scale factor is regressed at training. Future work will also involve transforming the scale regression of the RCNN into a multi direction scale factor transformation thus providing further information on the shape of the enclosed object.

  \bibliography{midl-samplebibliography}






\end{document}